\pdfoutput=1

\documentclass{article}

\usepackage[numbers]{natbib}

\usepackage[preprint]{neurips_2024}

\usepackage[utf8]{inputenc} 
\usepackage[T1]{fontenc}    
\usepackage{hyperref}       
\usepackage{url}            
\usepackage{booktabs}       
\usepackage{amsfonts}       
\usepackage{nicefrac}       
\usepackage{microtype}      
\usepackage{xcolor}         

\usepackage{amsmath}
\usepackage{multirow}

\usepackage[utf8]{inputenc} 
\usepackage[T1]{fontenc}    
\usepackage{lipsum}
\usepackage{graphicx}
\usepackage{amsmath}
\usepackage{multirow}
\usepackage{appendix}

\title{GFlowNet Pretraining with Inexpensive Rewards}

\author{
  Mohit~Pandey\thanks{Work done as a part of research internship at Recusion Pharmaceuticals } \\
  Vancouver Prostate Centre\\
  The University of British Columbia\\
  \texttt{mkpandey@student.ubc.ca} \\
  \And
  Gopeshh Subbaraj \\
  Mila - Quebec AI Institute \\
  Université de Montréal \\
  \texttt{gopeshh.subbaraj@mila.quebec} \\
  \AND
  Emmanuel Bengio \\
  Recursion Pharmaceuticals \\
  \texttt{emmanuel.bengio@recursionpharma.com} \\
}

\begin{document}

\maketitle

\begin{abstract}
Generative Flow Networks (GFlowNets), a class of generative models have recently emerged as a suitable framework for generating diverse and high-quality molecular structures by learning from unnormalized reward distributions. Previous works in this direction often restrict exploration by using predefined molecular fragments as building blocks, limiting the chemical space that can be accessed. In this work, we introduce Atomic GFlowNets (A-GFNs), a foundational generative model leveraging individual atoms as building blocks to explore drug-like chemical space more comprehensively. We propose an unsupervised pre-training approach using offline drug-like molecule datasets, which conditions A-GFNs on inexpensive yet informative molecular descriptors such as drug-likeliness, topological polar surface area, and synthetic accessibility scores. These properties serve as proxy rewards, guiding A-GFNs towards regions of chemical space that exhibit desirable pharmacological properties. We further our method by implementing a goal-conditioned fine-tuning process, which adapts A-GFNs to optimize for specific target properties. In this work, we pretrain A-GFN on the ZINC15 offline dataset and employ robust evaluation metrics to show the effectiveness of our approach when compared to other relevant baseline methods in drug design.
\end{abstract}

\vspace{-0.15in}
\section{Introduction}
\vspace{-0.08in}

GFlowNets are amortized samplers that learn stochastic policies to sequentially generate compositional objects from a given
unnormalized reward distribution. They can generate diverse sets of high-reward
objects, which has a demonstrated utility in small molecules drug-discovery tasks \citep{bengio2021flow}. Traditionally, GFlowNets for molecules generation have focused on fragment-based drug discovery i.e. the GFlowNet action space comprises of some predetermined fragments as building blocks for molecules. While producing diverse candidates, fragment-based approach limits the pockets of chemical space assessible by GFlowNet policy \citep{zhu2024sample,bengio2021flow,jain2023multi,shen2024tacogfn}. A truly explorative generative policy, tapping into the full potential of GFlowNet would be realizable when using atoms instead of fragments as the action space \citep{mouchlis2021advances,meyers2021novo,nishibata1991automatic}. On the other hand, the vastness of accessible state space makes training atom-based GFlowNets susceptible to collapse. Earlier atom-based GFlowNets attempts overcame this issue by limiting to small trajectories \citep{jain2023multi}.  However, since most commercially available drugs have molecular weights ranging from 200 to 600 daltons\citep{bos2000500,bickerton2012quantifying}, the molecules generated from these small trajectories are unlikely to possess the drug-like characteristics necessary for therapeutic efficacy.  In this work, we propose to mitigate the small trajectory length constraint for atom-based GFNs by pretraining them with expert demonstrations, coming in the form of offline drug-like molecules. Our main contributions in this paper are as following: \vspace{-0.08in}
\begin{itemize}
\item We introduce A-GFN-an atom-based GFlowNet for sampling molecules proportional to rewards governed by inexpensive molecular properties. 
\item We propose a novel strategy for unsupervised pretraining of A-GFNs by leveraging drug-like molecules using offline, off-policy training. This pretraining enables broader exploration of chemical space while maintaining diversity and novelty relative to existing drug-like molecule datasets.
    \item Through extensive experimentation, we demonstrate that goal-conditioned fine-tuning of A-GFNs for sampling molecules with desired properties offers significant computational advantages over training A-GFNs from scratch for the same objectives. 
    %
\end{itemize}
\vspace{-0.2in}
\section{Related Works}
\vspace{-0.1in}

\label{sec:headings}
\subsection{GFlowNet}
GFlowNets  were introduced by Bengio et al. in 2021 \citep{bengio2021flow} as a framework for training energy-based generative models that learn to sample diverse candidates in proportion to a given reward function. 
Unlike traditional Reinforcement Learning (RL) methods, which focus on maximizing rewards through a sequence of actions, GFlowNets aim to generate samples with probabilities proportional to their associated rewards. This distinction allows GFlowNets to explore a broader solution space, facilitating the discovery of novel, high-quality, and diverse objects across various domains. Recent works have been conducted along a multitude of directions, some focusing on theoretical aspects, such as connections to variational methods ~\citep{pmlr-v162-zhang22v, malkin2023gflownetsvariationalinference}), and some focused on improved training methods for better credit assignment and sample efficiency ~\citep{malkin2022trajectory, JMLR:v24:22-0364}. Due to its flexibility, GFlowNets have also been applied successfully to different settings such as biological sequences\citep{pmlr-v162-jain22a}, causal discovery ~\citep{pmlr-v180-deleu22a, atanackovic2023dyngfnbayesianinferencegene}, discrete latent variable modeling\citep{hu2023gflownetemlearningcompositionallatent}, and computational graph scheduling \citep{zhang2023robustschedulinggflownets}.

\label{sec:others}
\vspace{-0.08in}
\subsection{Unsupervised pretraining in RL and GFlowNets}
\vspace{-0.08in}
Pretraining models in machine learning has virtually become the default way to obtain powerful models~\citep{brown2020language,kalapos2024cnn}. Specifically, unsupervised pretraining in reinforcement learning (RL) has emerged as a promising strategy to enhance data efficiency and improve agent performance across various tasks. Recent works have explored different methodologies to leverage unsupervised interactions with the environment before fine-tuning on specific objectives. For instance, Liu and Abbeel (2020) introduced Active Pre-Training \citep{liu2021behaviorvoidunsupervisedactive}, a reward-free pre-training method that maximizes particle-based entropy in a contrastive representation space, achieving human-level performance on several Atari games and significantly enhancing data efficiency in the DMControl suite. Similarly, Mutti et al. \citep{mutti2021unsupervisedreinforcementlearningmultiple} addressed unsupervised RL in multiple environments, proposing a framework that allows for pre-training across diverse scenarios to improve the agent's adaptability. Additionally, the Unsupervised-to-Online RL framework \citep{kim2024unsupervisedtoonlinereinforcementlearning} was developed, which replaces domain-specific offline RL with unsupervised pre-training, demonstrating that a single pre-trained model can be effectively reused for multiple downstream tasks, often outperforming traditional methods. Similarly, In the context of GFlowNets, recent advancements have introduced unsupervised pre-training strategies, such as the outcome-conditioned GFlowNet \citep{pan2023pre}, which enables reward-free pre-training by framing the task as a self-supervised problem. This approach allows GFlowNets to learn to explore the candidate space and adapt efficiently to downstream tasks, showcasing the potential of unsupervised pre-training in enhancing the performance of generative models in molecular design.


\vspace{-0.08in}
\subsection{Goal-Conditioned and Multi-Objective Gflownets}
\vspace{-0.08in}
Recent advancements in GFlowNets have focused on goal-conditioned and multi-objective frameworks, enhancing their applicability in complex generative tasks. Goal-conditioned GFlowNets enable the generation of diverse outputs tailored to specific objectives, improving sample efficiency and generalization across different goals, as demonstrated by \citep{pmlr-v162-jain22a} and further refined through methods like Retrospective Backward Synthesis \citep{he2024lookingbackwardretrospectivebackward} to address sparse reward challenges. Roy et al., \citep{roy2023goalconditionedgflownetscontrollablemultiobjective}, impose hard constraints on a GFlowNet model by employing focus regions as a goal-design strategy which makes it comparable to a form of goal-conditional reinforcement learning \citep{schaul2015universal}. Additionally, the development of multi-objective GFlowNets \citep{jain2023multi} allows for simultaneous optimization of multiple criteria, providing practitioners with greater control over the generative process and the ability to explore trade-offs between competing objectives.

\vspace{-0.08in}
\subsection{Molecule Generation}
\vspace{-0.08in}
To contextualize our approach, we provide a brief review of prior research that utilized atom-based vocabularies in molecular generative modeling. Reinforcement learning (RL) studies optimal decision-making methodologies to maximize cumulative rewards. Given the shared notion of object-constructing Markov decision policies between GFlowNets and RL, we focus on the latter literature for molecule generation. For a comprehensive review of methods in molecule generation, we point the readers to \citep{bilodeau2022generative}. Recently, RL has been frequently employed in de novo design tasks due to its capability to explore chemical spaces beyond the compounds present in existing datasets. Moreover, it allows for targeted molecule generation for constrained property optimization. Early works in this domain focused on the auto-regressive generation of SMILES within an RL-loop for molecular property optimization\citep{olivecrona2017molecular,popova2018deep,goel2021molegular}. MolDQN\citep{zhou2019optimization} employs deep Q-Networks and multiobjective molecular properties for scalarization of rewards to generate 100\% valid molecules without pretraining on a dataset. 
You et al.\citep{you2018graph} and Atance et al.\citep{atance2022novo} have employed graph neural networks, trained on offline molecular datasets, to generate molecular graphs. They simultaneously applied policy-gradient reinforcement learning to ensure that the generated molecules adhere to specified property profiles.




\vspace{-0.15in}
\section{Preliminaries}
\vspace{-0.1in}

GFlowNets are generative models designed to sample structured objects from a state space $\mathcal{S}$ in proportion to a reward function $R\left(s_T\right)$ assigned to terminal states $s_T$. The model generates trajectories $\tau=\left(s_0, a_0, \ldots, s_T\right)$ by transitioning between states through actions, guided by a forward policy $P_F\left(s^{\prime} \mid s\right)$. The key idea is to learn a flow function $F(s)$ that ensures the total flow into each state equals the flow out, so the probability of reaching any terminal state is proportional to its reward. This is achieved by aligning the forward and backward policies $P_F$ and $P_B$, ensuring the final distribution over terminal states reflects the desired reward distribution.

Existing extensions of GFlowNets that handle multiple, potentially conflicting objectives \citep{jain2023multi,zhu2024sample}, along with the benefits of scalability to large action spaces and credit assignment makes them well-suited for molecular generation tasks; the primary focus of our work. We consider molecules as their topological graphs $G = (A,E,\mathcal{X})$ such that $\mathcal{X} \in \mathbb{R}^{n \times d}$ define d-dimensional atomic features for the $n$ nodes in $G$, $E \in \{0,1\}^{b\times n \times n}$ is edge-adjacency tensor for $b$ types of edges connecting $n$ nodes and finally $A \in \{0,1\}^{n \times n}$ is the adjacency matrix for $n$ nodes. The corresponding trajectory $\tau$ for $G$ $( \tau = (s_0,a_0),....(s_n, a_n))$ is generated by sampling actions $a$ from a conditional learning policies $P_F(.\mid c; \theta)$ and $P_B(.\mid c,G;\theta)$ of the GFlowNet $\mathcal{G_\theta}$, where state $s_i$ is a partially constructed subgraph of $G$, and $s_n$ = $G$.

Our primary objective is to learn $\pi$ such that the generated $G$ are chemically valid molecular graphs satisfying some defined molecular property conditional ranges $c$, and exhibiting sufficient diversity. Specifically, we want to train a conditional policy which samples molecular graphs $G$ with probability proportional to $R(G|c)$, where $R(G|c)$ is a reward function measuring how well generated $G$ satisfies $c$. Our secondary objective involves fine-tuning this pretrained $\mathcal{G}_\theta = (P_F(;\theta), P_B(;\theta))$  to achieve drug discovery tasks with certain molecular property constraints and establish the benefits of fine-tuning GFlowNets. Trajectory balance \citep{malkin2022trajectory} being the most common learning objective designed to improve credit assignment in GFlowNets is used as the training method in this work. 
\vspace{-0.08in}
\begin{equation}\label{TBeq}
\mathcal{L}_{\text{TB}}(\tau) = \left( \log \left( \frac{Z_\theta \prod_{t=1}^{n} P_F(s_t | s_{t-1}; \theta)}{R(x) \prod_{t=1}^{n} P_B(s_{t-1} | s_t; \theta)} \right) \right)^2
\end{equation}
where, trajectory $\tau = (s_0, s_1,....,s_n)$ such that $s_n=x$ is a fully constructed object. Within the context of our work, $x \in \mathcal{X}$ are samples from the combinatorial space of all possible molecules using actions $a_t \sim \mathcal{A}$. 
This equation ensures that the product of forward policy probabilities along a trajectory is proportional to the product of backward policy probabilities and the reward along the same trajectory.


\vspace{-0.15in}
\section{Unsupervised Pretraining with Inexpensive Rewards}
\vspace{-0.1in}


To construct molecular graphs, we design an action space with 5 action types. The agent can add a node (heavy atom) to the graph, add an edge between two nodes, set a node's properties (e.g. its chirality), set a bond's properties (i.e. its bond order), or stop the trajectory. We use a graph neural network~\citep{velivckovic2017graph} 
to parameterize a policy with such actions, using the GNN's invariances to produce per-node and per-edge logits. We are also careful to mask these logits such that the produced molecules always have valid valences (i.e. by design the molecules are always convertible to RDKit molecules and SMILES). Since our method generates molecules atom by atom, we refer to the approach as \textbf{Atomic GFlowNet (A-GFN)}.
\vspace{-0.08in}
\subsection{Inexpensive Molecular Rewards} \label{rewards} 
\vspace{-0.08in}
In the context of pre-training A-GFN models for molecular design, we utilize inexpensive molecular rewards—such as Topological Polar Surface Area (TPSA), Quantitative Estimate of Drug-likeness (QED), synthetic accessibility (SAS), and the number of five or six-membered rings in a molecule. These rewards are computationally cheap to evaluate and serve as proxies for more complex properties. Fine-tuning the models is then conducted on more expensive and computationally intensive tasks, such as predicting binding affinity or toxicity (e.g., LD50), which are crucial for drug discovery but require significant computational resources or experimental data to assess accurately. 
\vspace{-0.08in}
\subsubsection{Reward Function}
\vspace{-0.08in}
In order to pre-train a goal-conditioned A-GFN for learning molecular properties 
$p\in P$, we need to define the property-specific conditional ranges $c_p = (c_{low},c_{high})$ from which molecules are generated. We use the following goal-conditioned reward function for property $p$ and molecule $x$:
\begin{equation}\label{eq:rew_d_pos}
   R_p(x | c_p, \Vec{d}>0) = \text{reward}(p_x\mid c_p, \Vec{d}>0) =
\begin{cases}
0.5*\exp \left( -\frac{(\text{c$_{low}$} - p_x)}{\lambda} \right) & \text{if } p_x < \text{c$_{low}$} 
\\
\exp \left( -\frac{(p_x - \text{c$_{high}$})}{\lambda} \right) & \text{if } p_x > \text{c$_{high}$} 
\\
\frac{0.5*(p_x - \text{c$_{low}$})}{(\text{c$_{high}$} - \text{c$_{low}$})} + 0.5 & \text{otherwise}
\end{cases}
\end{equation}
here, $\lambda$ controls the decay rate, $c_{low}$ and $c_{high}$
  are the lower and upper bounds for property $p$, and $\Vec{d} \in \mathbb{R}$ represents the $preference\_direction$ hyperparameter, indicating whether lower or higher property values within the range $c_p$ are preferred (for further details, see appendix \ref{app:cond_and_rew},\ref{app:metrics}). Note that during training, we sample these ranges so that the model learns to be robust to inference-time queries (sec \ref{condinfo}).
  
Drug discovery is inherently a multiobjective optimization problem where the drug candidates are expected to simultaneously have several desired properties, such as a high drug-likeliness (QED), high SAS, and reasonably low TPSA, among other criteria. In our de novo molecular generation setup, we wish to satisfy the same multiobjective desiderata \citep{jain2023multi}. We choose this aggregated scalarization of the multiobjective reward over property set $P$ as
\begin{equation}\label{rewpretraineq}
R(x|c_{p_1}, ..., c_{p_{|P|}}) = \prod_{p\in P}R_p(x|c_p)     
\end{equation}
\vspace{-0.08in}
\subsubsection{Reward conditioning A-GFN}\label{condinfo} 
\vspace{-0.08in}
To alleviate the problem of sparse rewards in training A-GFNs for molecular graph generation with some hard constraints, we condition the sampler by using a distribution of goals derived from reasonable lower and upper bounds of the molecular property ranges we care about. This conditioning effectively narrows down the search space to regions of interest defined by these property ranges, ensuring that the generated molecules are not only diverse but also relevant to the specific objectives of the drug discovery process.
In order to ensure that the A-GFN does not develop a selective bias towards specific values within a property range and instead explores the full range of possible values for each molecular property $p$, we sample the conditional vectors $c_{p,j}$ uniformly across their respective ranges.
Specifically, for the $j^{th}$ online trajectory in the batch, the lower and upper bounds $c_{p,j}^{low}, c_{p,j}^{high}$ for the property $p$ are drawn from a uniform distribution as $c_{p,j}^{low}, c_{p,j}^{high} \sim U(c_{low},c_{high})$, where $c_{low}$ and $c_{high}$ are the predefined desired lower and upper bounds for the property $p$. 
In cases where the values for properties $p$ for the trajectory $j$ leading to a valid molecular graph are known \textit{a priori} ($p_j$) (e.g., from molecules in an offline dataset), $c_{p,j}$ are centered around these known values, i.e. $c_{p,j}^{low}, c_{p,j}^{high} \sim  \mathcal{T}(p_j, \sigma_p, c_{low}, c_{high})$, where $\mathcal{T}$ is a truncated normal distribution centered at the value of property $p$ calculated for trajectory $j$, $\sigma_p$ is a hyperparmeter controlling the variance of $\mathcal{T}$ for property $p$.

Such a probabilistic goal-sampling strategy ensures that each trajectory within the batch is conditioned on a randomly selected sub-range within the broader property range. This enables our laid out objective of promoting exploration across the entire desired chemical space and preventing the A-GFN from overfitting to narrow regions within the chemical space. In order to further prevent A-GFN from memorizing the specified property ranges but rather encourage it to learn to sample molecules within any arbitrary range, we provide negative samples by sampling out-of-bounds conditionals with a small probability. To this end, with $\epsilon \approx 0$, we sometimes sample $c_{p,j}^{low}, c_{p,j}^{high}$ from $\mathcal{U}(c_p^{low},c_p^{high})$ and $\mathcal{U}(c_{p,j}^{low},c_{p}^{high})$ respectively for the online trajectories, while for offline trajectories, these conditional bounds are sampled as following 
\[
(c_{p,j}^{low}, c_{p,j}^{high}) = 
\begin{cases}
(c_p^{low}*, \mathcal{U}(c_p^{low}*, p_j)) & \text{if } B_i = 0 \\
(\mathcal{U}(p_j,c_p^{high}*), c_p^{high}*) & \text{if } B_i = 1
\end{cases}
\]
where  $B_i$  is a Bernoulli random variable with \( P(B_i = 0)= P(B_i = 1) = 0.5 \) and $c_p^{low}*$ and $c_p^{high}*$ are minimum and maximum permissible values for molecular property $p$. 
Finally, $c_{p,j}$ is constructed as a vector by applying thermometer encoding \citep{buckman2018thermometer} to the sampled scalar bounds $c_{p,j}^{low}$ and $c_{p,j}^{high}$.

\vspace{-0.08in}
\subsection{Pretraining GFN with expert offline trajectories}\label{sec:offln_trajs}
\vspace{-0.08in}
Bengio et al. \citep{bengio2021flow} employ replay buffer-based off-policy training for GFlowNets to generate novel molecules. However, such online off-policy methods can suffer from high variance \citep{fedus2020revisiting,vemgal2023empirical} and and a lock-in of suboptimal trajectories, particularly in sparse reward settings, where the agent struggles to adequately explore rare, high-reward regions of the state space, resulting in slow convergence and suboptimal performance. To mitigate these challenges, we propose leveraging the vast amounts of readily available inexpensive and unlabelled molecular data to perform a hybrid online-offline off-policy pretraining of A-GFN. This data provides valuable expert trajectories and the molecular properties derived from this data can provide inexpensive extrinsic rewards, giving a better starting point for exploration. We form our training batches by integrating offline expert trajectories from the ZINC dataset ($\mathcal{D}_{ZINC}$) with online updates  ($\tau = \tau_{online} \oplus \tau_{offline}$).  $\tau_{offline}$ are generated from these molecules, $x\in \mathcal{D}_{ZINC}$ by sampling deleterious actions \{$\texttt{deleteNode}$, $\texttt{deleteEdge}$, $\texttt{removeNodeAttribute}$, $\texttt{removeEdgeAttribute}$\} according to a conditional backward policy $P_B$ as $\tau_{offline}$ $\sim$ $P_B(.\mid x, c_p;\theta)$. Likewise, online trajectories are created by sampling constructive actions \{$\texttt{addNode}$, $\texttt{addEdge}$, $\texttt{addNodeAttribute}$, $\texttt{addEdgeAttribute}$, $\texttt{stop}$\} from conditional forward policy P$_F$ as $\tau_{online} \sim P_F(.\mid c_p;\theta).$

\vspace{-0.08in}
\subsection{Regularized Loss Balancing for Exploration and Synthesis Feasibility}
\vspace{-0.08in}
The primary allure of molecular generative models with large explorative capacities such as A-GFN is their ability to navigate the near-infinite possibilities of chemical structures. On the other hand, ensuring that generated molecules lie close to the chemical space feasible for synthesis,   particularly within the constraints of make-on-demand (MOD) libraries such as ZINC, is crucial for synthesis and \textit{in vitro} validation in drug discovery. To balance these two seemingly conflicting goals, we introduce a regularization term in the pretraining objective. Specifically, we combine the exploration objective with a Maximum Likelihood Estimation (MLE) loss over the offline dataset, leading to the following regularized loss function:
\vspace{-0.08in}
\begin{equation}\label{eq:gfn_mle_loss}
\mathcal{L}= \lambda_1\mathcal{L}_{TB}+\lambda_2\mathcal{L}_{MLE} 
\end{equation} 
$$\textit{where, }  \mathcal{L}_{MLE} = -\log(P_F(.\mid x, c_p)) \forall x \in D_{ZINC}$$
$\mathcal{L}_{TB}$ encourages exploration of the chemical space, $\mathcal{L}_{MLE}$ is the distributional learning loss term ensuring proximity to MOD libraries, and $\lambda_1, \lambda_2$ are hyperparameters controlling the trade-off between exploration and adherence to the MOD space (Tab.\ref{tab:hps}). 
With the online and offline trajectories described in sec\ref{sec:offln_trajs} and molecular property rewards in sec\ref{rewards}, we optimize the prior A-GFN ($\mathcal{G}_\theta$) by minimizing Eq.\ref{eq:gfn_mle_loss} until convergence in reward.



\vspace{-0.15in}
\section{Finetuning}
\vspace{-0.1in}
In this section, we investigate the methodology for utilizing the pre-trained A-GFN model ($\mathcal{G}_\theta$) and its subsequent adaptation to downstream drug discovery tasks based on harder reward functions. To fine-tune the model, the pre-trained parameters of $\mathcal{G}_\theta$ serve as the initialization for task-specific adaptation. This initialization enables the model to retain useful structural priors from pretraining, thus improving sample efficiency and convergence speed during fine-tuning. In particular, we retrain $\mathcal{G}_\theta$ by integrating task-specific reward $R_{ext}$. This modifies eq.\ref{rewpretraineq} as
\begin{equation}\label{eq:rext}
R(x|c_{p_1}, ..., c_{p_{|P|}}) = \prod_{p\in P}R_p(x) \times R_{ext}(x)    
\end{equation}
Such a reward formulation ensures that the GFlowNet receives a high reward for generating molecules that simultaneously follow the desired molecular properties and are highly suitable for the downstream task. It should be noted that $Z_\theta$ in eq \ref{TBeq} is a global scalar that estimates the normalization constant for the unnormalized reward function $R(x)$ \big(i.e, $Z_\theta = \sum_{x\in\mathcal{X}} R(x)$\big). Thus, to enable $Z_\theta$ to generalize to the new $R(x)$, we inject noise into the pretrained A-GFN model's parameters, a common strategy in pretrain-then-finetune approaches \citep{yuan-etal-2023-hype, tong2022robust}.

\vspace{-0.15in}
\section{Experiments}
\vspace{-0.1in}
We evaluate A-GFN's effectiveness during both pretraining and fine-tuning using comprehensive metrics such as novelty, diversity, uniqueness, success rate, validity, L1-distance, and number of modes. Full details are provided in appendix \ref{app:metrics}.

For pretraining, we compare the performance of A-GFN against fragment-based GFlowNet, conditioned on the same molecular properties. Our results show that A-GFN significantly outperforms fragment-GFN in exploring drug-like chemical space across multiple objectives. For fine-tuning, we benchmark the pre-trained A-GFN against A-GFN trained from scratch (task-trained A-GFN) on several downstream drug discovery tasks. Following the task setup in \citep{you2018graph}, we focus on the following finetuning objectives:\\
\textbf{Property Optimization}: The goal here is to generate molecules that maximize or minimize a specified physicochemical, structural, or binding property while ensuring diversity among the generated molecules. In this unconstrained optimization setting, we compare the fine-tuned A-GFN to the task-trained A-GFN and other state-of-the-art methods for unconstrained molecule generation. For fairness, we only include atom-based generative methods in the baselines, excluding fragment-based approaches.\\
\textbf{Property Targeting}: In this task, the objective is to generate molecules that adhere to predefined molecular property ranges while being structurally distinct from the training (pretraining) set.\\
\textbf{Property Constrained Optimization}: This task requires the generation of molecules that simultaneously meet both property optimization and property targeting criteria i.e. molecules must lie within the specified property ranges while also maximizing or minimizing the targeted property.

\vspace{-0.1in}
\subsection{Pretraining}\label{subsec:pretraining}
\begin{figure}
\includegraphics[width=\textwidth,height=\textheight,keepaspectratio]{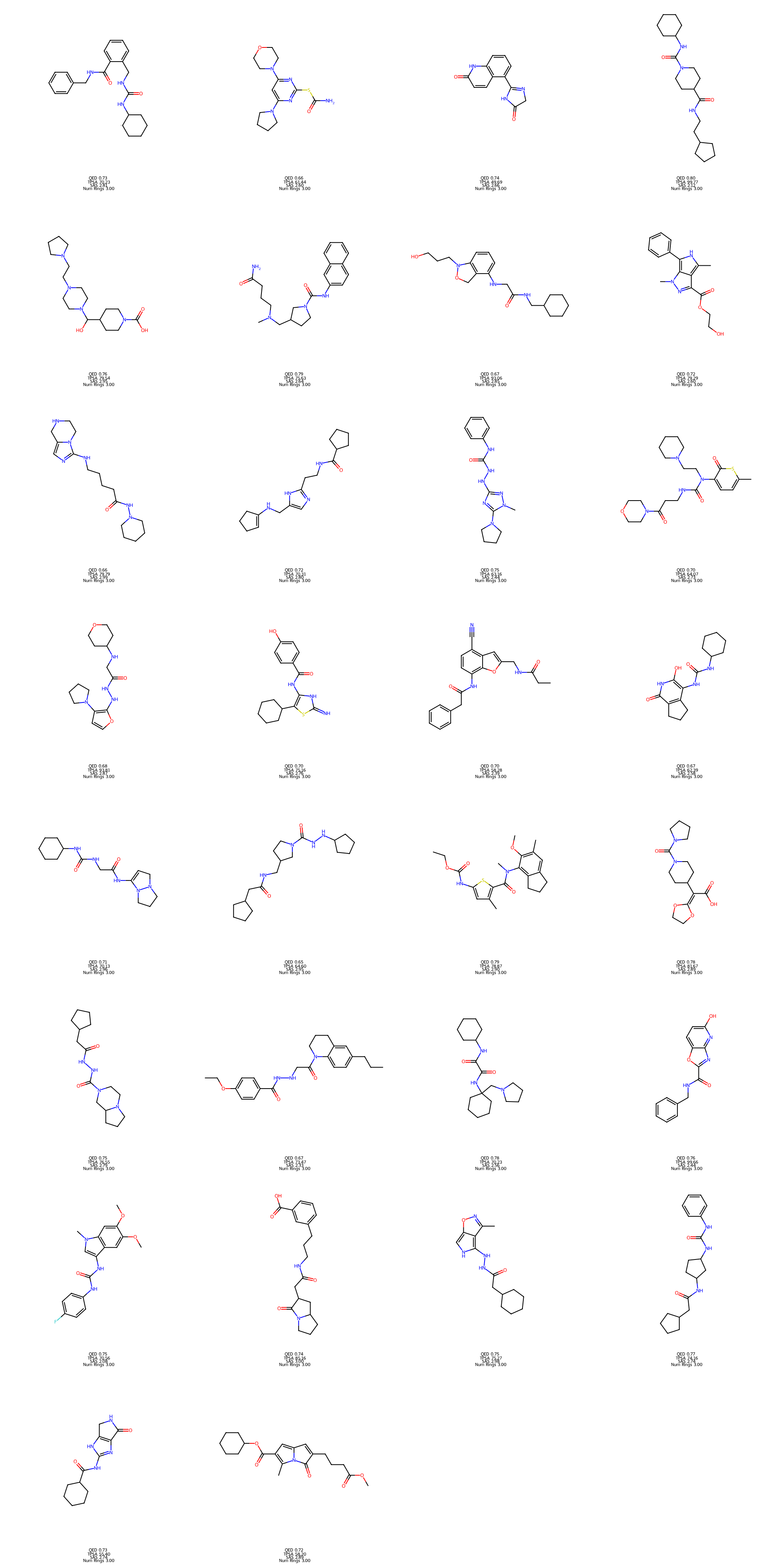} 
  \centering
  \caption{Some randomly chosen successful molecules from the pretrained A-GFN.}
  \label{fig:pretrain_mols}
\end{figure}

\vspace{-0.1in}
\textbf{A-GFN pretraining setup}: 
To create a versatile foundation for a range of downstream molecular generation tasks, we train our A-GFN using a hybrid online-offline off-policy strategy. For our pretraining, we utilize ZINC250K, a curated subset of the ZINC database, which consists of 250,000 commercially accessible drug-like compounds drawn from over 37 billion molecules available in ZINC \citep{tingle2023zinc}. The primary goal during pretraining is to optimize the A-GFN for generic yet critical drug-like properties: TPSA, QED, SAS, and the number of rings. The desired ranges for these properties are enumerated in Tab.\ref{tab:task_des_range}. These properties are chosen based on their established relevance in guiding molecular design towards compounds with desirable pharmacokinetic and pharmacodynamic profiles, following the framework of \citep{wellnitz2024stoplight}. By optimizing for these properties, we aim to equip A-GFN with the capability to generate molecules that strike a balance between drug-likeness and structural diversity. We restrict atom types to a core set commonly found in drug-like molecules: $C, S, P, N, O, F$, and implicit hydrogen ($H$). This ensures that the generated molecules are synthetically relevant and pharmacologically plausible, avoiding rare or exotic atom types that are less likely to lead to viable drug candidates. For benchmarking fragment-based GFlowNets, we adapt the purely online training setup proposed in \citep{bengio2021flow}, conditioning it on the same molecular properties as A-GFN. To align with our A-GFN setup, we generate a fragment vocabulary from the BRICS decomposition of ZINC250K, selecting the 73 (following \citep{bengio2021flow}) most common fragments. This equips the fragment-based GFlowNet with a diverse and representative set of building blocks for molecular generation. Through pretraining, we observe that A-GFN effectively adapts to the specified property ranges while maintaining high molecular diversity and uniqueness. All molecules generated by A-GFN are valid, adhering to chemical rules, and novel, as they do not replicate any molecules in the ZINC250K dataset (Fig.\ref{fig:pretrain_mols}). For the same set of property conditionals, A-GFN demonstrates superior chemical scaffold exploration compared to the fragment-based GFlowNet, covering nearly twice as many distinct scaffolds (Tab.\ref{tab:pretraining}). While the fragment-based method has a higher success rate and better control over specific molecular properties, A-GFN excels in uniqueness and novelty, making it a more powerful tool for exploring uncharted chemical space in drug discovery. This highlights the model's capacity to explore novel regions of chemical space while adhering to fundamental molecular design principles.

\begin{table}[]
\caption{Comparing pretraining A-GFN to pertaining a fragment-based GFlowNet for same molecular property conditionals. Pretraining was performed on Nvidia A100-40G GPUs.}
\label{tab:pretraining}
\renewcommand{\arraystretch}{1.5}
\resizebox{\textwidth}{!}{
\begin{tabular}{|c|c|c|c|cccc|c|c|c|c|c|}
\hline
\textbf{Method}                                             & \textbf{N\_modes} & \textbf{Diversity} & \textbf{Success \%} & \multicolumn{4}{c|}{\textbf{L1-dist ($\downarrow$)}}                                                                                    & \textbf{Uniqueness} & \textbf{Novelty} & \textbf{Validity} & \textbf{\begin{tabular}[c]{@{}c@{}}Scaffolds\\ (n=200)\end{tabular}} & \textbf{\begin{tabular}[c]{@{}c@{}}Time\\ (GPU-hours)\end{tabular}} \\ \hline
                                                            &                   &                    &                     & \multicolumn{1}{c|}{TPSA} & \multicolumn{1}{c|}{\begin{tabular}[c]{@{}c@{}}Num\\ Rings\end{tabular}} & \multicolumn{1}{c|}{SAS}  & QED  &                     &                  &                   &                                                                      &                                                                     \\ \hline
\begin{tabular}[c]{@{}c@{}}Fragment\\ GFlowNet\end{tabular} & 0                 & 0.002              & \textbf{50.00 }              & \multicolumn{1}{c|}{0.00} & \multicolumn{1}{c|}{1.50}                                                & \multicolumn{1}{c|}{0.00} & 0.71 & 0.001               & 0.97             & 0.99              & 113                                                                  & \multirow{2}{*}{514.0}                                                \\ \cline{1-12}
A-GFN                                                       & \textbf{1252 }             &\textbf{ 0.88 }              & 45.09               & \multicolumn{1}{c|}{0.03} & \multicolumn{1}{c|}{0.49}                                                & \multicolumn{1}{c|}{0.24} & 0.67 & \textbf{0.96}                &\textbf{ 1.0}              & 1.0               & \textbf{196}                                                                  &                                                                     \\ \hline
\end{tabular}
}
\vspace{-0.2in}
\end{table}

\vspace{-0.15in}
\subsection{Fine-tuning}
\vspace{-0.08in}

\textbf{Fine-tuning setup} We split the fine-tuning tasks' objectives as property optimization, property targeting, and property constrained optimization. The primary distinction between these three setups is in terms of the corresponding conditionals and reward function. For the tasks where some small labeled dataset is accessible, we investigate the benefits of offline data on fine-tuning atomic-GFlowNets. Similar to GCPN \citep{you2018graph}, we consider two structural tasks, molecular weight (mol.wt.), and logP (partition coefficient for drug's water to octanol concentration \citep{sangster1997octanol}). In addition, we also consider other standard drug-discovery tasks where rewards are based on empirical models\citep{huang2021therapeutics} such as the LD50 (toxicity) task of \citep{zhu2009quantitative}.  For each task and objective pair, we aim to show that fine-tuning a pretrained GFlowNet achieves the task's objectives more quickly than training a new GFlowNet from scratch.  We now define each objective in detail.
\vspace{-0.15in}
\subsubsection{Property Optimization}
\vspace{-0.08in}
In this task, the goal is to generate novel molecules that minimize specific physicochemical properties. While previous works have commonly benchmarked their models on QED optimization \citep{jeon2020autonomous,zhou2019optimization,you2018graph}, we exclude QED from our evaluation, as our pretrained A-GFN is already optimized for this metric. Instead, we focus on mol.wt. and logP—two critical properties in drug discovery. To further challenge the models, we set the target property ranges to be slightly below the minimum values found in the Zinc250K dataset, with logP in the range [-5, -4.5] and molecular weight in the range [100, 110]. The success percentage is calculated based on the proportion of generated molecules that fall within these strict property ranges. The low success percentages across all methods reflect the challenging nature of the task (Tab.\ref{tab:prop_opt_baselines}). However, the fine-tuned A-GFN shows a relatively higher success rate compared to other methods, combined with a lower-L1 distance from the target ranges. This indicates that even in cases where exact matches are not achieved, the fine-tuned A-GFN generates molecules that are close to the desired property values. A-GFN’s ability to sample from these "out-of-domain" chemical spaces—where molecules fall outside the property ranges in MOD datasets like Zinc250K—suggests that fine-tuned A-GFN excels in generating diverse molecules that could extend the scope of current chemical libraries. Moreover, the high diversity, novelty, and validity of the molecules generated by A-GFN indicate its capacity to produce a broad spectrum of unique, structurally valid compounds, essential for mitigating structural redundancy in drug candidates. Fine-tuning further enhances these aspects, highlighting A-GFN’s potential for property-driven molecular design.

\begin{table}[]
\caption{Comparing fine-tuned A-GFN with other baselines for logP and molecular weight property optimization task.}

\label{tab:prop_opt_baselines}
\renewcommand{\arraystretch}{1.5}
\resizebox{\textwidth}{!}{
\begin{tabular}{|c|c|c|c|c|c|c|c|c|c|}
\hline
\textbf{Task}                                                               & \textbf{Method}                                              & \textbf{N\_modes} & \textbf{Diversity} & \textbf{Success \%} & \textbf{\begin{tabular}[c]{@{}c@{}}L1-dist ($\downarrow$)\\ task\end{tabular}} & \textbf{Uniqueness} & \textbf{Novelty} & \textbf{Validity} & \textbf{\begin{tabular}[c]{@{}c@{}}Time\\ (GPU-hours)\end{tabular}} \\ \hline
\multirow{4}{*}{logP}                                                       & GCPN                                                         & 154               & \textbf{0.91}      & 2.54                & 0.45                                                                           & \textbf{1.0}        & \textbf{1.0}     & 1.0               & 40.0                                                                \\ \cline{2-10} 
                                                                            & Reinvent                                                     & 28                & 0.90               & 0.34                & 0.41                                                                           & 0.93                & 0.99             & 0.99              & 4.0                                                                 \\ \cline{2-10} 
                                                                            & \begin{tabular}[c]{@{}c@{}}A-GFN\\ Task trained\end{tabular} & 928               & 0.89               & 13.02               & 0.17                                                                           & \textbf{1.0}        & \textbf{1.0}     & 1.0               & 20.0                                                                \\ \cline{2-10} 
                                                                            & \begin{tabular}[c]{@{}c@{}}A-GFN\\ Finetuned\end{tabular}    & \textbf{1199}     & 0.87               & \textbf{16.67}      & \textbf{0.19}                                                                  & 0.99                & \textbf{1.0}     & 1.0               & 20.0                                                                \\ \hline
\multirow{4}{*}{\begin{tabular}[c]{@{}c@{}}Molecular\\ Weight\end{tabular}} & GCPN                                                         & 223               & 0.92               & 4.02                & 0.14                                                                           & \textbf{1.0}        & \textbf{1.0}     & 1.0               & 40.0                                                                \\ \cline{2-10} 
                                                                            & Reinvent                                                     & 1                 & 0.88               & 0.02                & 0.33                                                                           & 0.97                & 0.99             & 0.99              & 4.0                                                                 \\ \cline{2-10} 
                                                                            & \begin{tabular}[c]{@{}c@{}}A-GFN\\ Task trained\end{tabular} & 9                 & 0.93               & 0.88                & 0.40                                                                           & \textbf{1.0}        & \textbf{1.0}     & 1.0               & 8.0                                                                 \\ \cline{2-10} 
                                                                            & \begin{tabular}[c]{@{}c@{}}A-GFN\\ Finetuned\end{tabular}    & \textbf{1157}     & \textbf{0.94}      & \textbf{6.58}       & \textbf{0.05}                                                                  & 0.95                & \textbf{1.0}     & 1.0               & 8.0                                                                 \\ \hline
\end{tabular}
}
\vspace{-0.2in}
\end{table}
\vspace{-0.15in}
\subsubsection{Property Targeting}
\vspace{-0.08in}
Here, we generate molecules that satisfy specific molecular property constraints, demonstrating the model's capacity for fine-tuning and adaptation to new objectives. Compared to training A-GFN from scratch, fine-tuning a pretrained A-GFN leads to significantly faster convergence, even when property ranges are altered. In particular, we focus on modifying the TPSA property while keeping the constraints for other molecular properties the same as those used in pretraining.

To evaluate the model's ability to generalize to new property constraints, we alter the TPSA range from its pretraining interval of [60, 100] to both lower [40, 60] and higher [100, 120] intervals. Our experiments demonstrate that the fine-tuned A-GFN consistently outperforms the model trained from scratch in terms of convergence speed and overall performance. This is particularly evident in the number of distinct high-reward molecular modes discovered (N\_modes), molecular diversity, and success percentage, all of which show substantial improvements after fine-tuning (Tab.\ref{tab:prop_tgt_opt}). The ability of the fine-tuned A-GFN to adapt to modified property constraints highlights the robustness of the pretrained $\mathcal{G}_\theta$ model, which has not merely memorized specific property ranges but has learned a more generalizable sampling strategy. This adaptability is crucial in real-world drug discovery applications, where molecular property requirements often shift during the drug discovery process.

\subsubsection{Property Constrained Optimization}
\vspace{-0.08in}
This comprehensive benchmarking task aims to generate molecules that minimize a target property (e.g., mol.wt., logP, or toxicity) within a predefined range, while maintaining the drug-like characteristics encoded during pretraining. We evaluate two key scenarios: conditionals-preserved fine-tuning and Dynamic Range adjustment (DRA). In the conditionals-preserved fine-tuning setup, we retain the same conditional property ranges ($c_p$) used during pretraining, setting the task’s $preference\_direction$ to -1, which directs the model to minimize the target property. The desired ranges for these task properties are set between the $25^{th}$ percentile of the ZINC dataset and a predetermined maximum threshold (see Tab.\ref{tab:pres_prop_const_opt} and \ref{tab:LD50_const_prop_opt} for specifics). The fine-tuned A-GFN significantly outperforms the A-GFN trained from scratch, achieving superior results within the same computational budget. In the dynamic range adjustment scenario, we modify one of the pretraining conditionals (TPSA) by shifting its range from $60 \leq TPSA \leq 100$ to both lower ($40 \leq TPSA \leq 60$) and higher ($100 \leq TPSA \leq 120$) values. In this case, fine-tuned A-GFN again surpasses its scratch-trained counterpart, demonstrating faster convergence and higher success rates (Tab.\ref{tab:der_prop_const_opt}). We also explore a hybrid online-offline fine-tuning approach, where A-GFN leverages offline task-specific data with the desired $c_p$, similar to its pretraining setup. In most tasks, hybrid fine-tuning shows comparable performance to fully online fine-tuning, 
with notable exceptions in more complex tasks like logP optimization when $100 \leq TPSA \leq 120$; it is plausible that in such a case grounding the model in data stabilizes early learning when rewards are low. In this challenging case, the goal was to generate molecules with logP$\approx$1.5 while maintaining the conditional properties outlined in Tab.\ref{tab:task_des_range}. The inherent difficulty of this task is underscored by the fact that only 0.002\% of molecules in the ZINC250K dataset meet these stringent criteria. Despite the challenge, A-GFN fine-tuned with a small offline dataset of expert trajectories achieve a respectable success rate and uncover diverse modes in the chemical space. This underscores the potential of hybrid online-offline fine-tuning to unlock otherwise inaccessible regions of the chemical landscape, offering a promising strategy for tackling difficult-to-sample molecular spaces.

\begin{table}[]
\caption{Comparing the effectiveness of finetuned A-GFN over A-GFN trained from scratch for conditional preserved finetuning.}
\label{tab:pres_prop_const_opt}
\renewcommand{\arraystretch}{1.5}

\resizebox{\textwidth}{!}{
\begin{tabular}{|c|c|c|c|c|ccccc|c|c|c|c|}
\hline
\textbf{Task}                                                                & \textbf{Method}                                                       & \textbf{N\_modes} & \textbf{Diversity} & \textbf{Success \%} & \multicolumn{5}{c|}{\textbf{L1-dist ($\downarrow$)}}                                                                                                                 & \textbf{Uniqueness} & \textbf{Novelty} & \textbf{Validity} & \textbf{\begin{tabular}[c]{@{}c@{}}Time \\ (GPU-hours)\end{tabular}} \\ \hline
                                                                             &                                                                       &                   &                    &                     & \multicolumn{1}{c|}{TPSA} & \multicolumn{1}{c|}{\begin{tabular}[c]{@{}c@{}}Num \\ Rings\end{tabular}} & \multicolumn{1}{c|}{SAS}  & \multicolumn{1}{c|}{QED}  & Task &                     &                  &                   &                                                                      \\ \hline
\multirow{3}{*}{\begin{tabular}[c]{@{}c@{}}Molecular \\ Weight\end{tabular}} & \begin{tabular}[c]{@{}c@{}}A-GFN\\ Task Trained\end{tabular}          & 0                 & \textbf{0.92}      & 15.71               & \multicolumn{1}{c|}{0.33} & \multicolumn{1}{c|}{1.50}                                                 & \multicolumn{1}{c|}{1.96} & \multicolumn{1}{c|}{2.32} & 0.14 & 0.95                & 1.0              & 1.0               & \multirow{2}{*}{24.0}                                                \\ \cline{2-13}
                                                                             & \begin{tabular}[c]{@{}c@{}}A-GFN\\ Finetuned\end{tabular}             & 340               & 0.71               & \textbf{43.09}      & \multicolumn{1}{c|}{0.06} & \multicolumn{1}{c|}{0.75}                                                 & \multicolumn{1}{c|}{0.12} & \multicolumn{1}{c|}{0.99} & 0.04 & 0.62                & 1.0              & 1.0               &                                                                      \\ \cline{2-14} 
                                                                             & \begin{tabular}[c]{@{}c@{}}A-GFN \\ Finetuned \\ w/ data\end{tabular} & \textbf{963}      & 0.86               & 35.52               & \multicolumn{1}{c|}{0.03} & \multicolumn{1}{c|}{0.39}                                                 & \multicolumn{1}{c|}{0.41} & \multicolumn{1}{c|}{0.94} & 0.08 & \textbf{1.0}        & 1.0              & 1.0               & 24.0                                                                 \\ \hline
\multirow{3}{*}{logP}                                                        & \begin{tabular}[c]{@{}c@{}}A-GFN\\ Task Trained\end{tabular}          & 0                 & \textbf{0.96}      & 20.32               & \multicolumn{1}{c|}{0.64} & \multicolumn{1}{c|}{1.18}                                                 & \multicolumn{1}{c|}{1.98} & \multicolumn{1}{c|}{2.59} & 0.48 & 0.75                & 1.0              & 1.0               & \multirow{2}{*}{1.0}                                                 \\ \cline{2-13}
                                                                             & \begin{tabular}[c]{@{}c@{}}A-GFN\\ Finetuned\end{tabular}             & \textbf{1286}     & 0.86               & 41.20               & \multicolumn{1}{c|}{0.04} & \multicolumn{1}{c|}{0.45}                                                 & \multicolumn{1}{c|}{0.31} & \multicolumn{1}{c|}{0.49} & 0.20 & \textbf{1.0}        & 1.0              & 1.0               &                                                                      \\ \cline{2-14} 
                                                                             & \begin{tabular}[c]{@{}c@{}}A-GFN\\ Finetuned\\ w/ data\end{tabular}   & 1080              & 0.87               & \textbf{43.47}      & \multicolumn{1}{c|}{0.05} & \multicolumn{1}{c|}{0.49}                                                 & \multicolumn{1}{c|}{0.28} & \multicolumn{1}{c|}{0.65} & 0.31 & 0.98                & 1.0              & 1.0               & 1.0                                                                  \\ \hline
\end{tabular}
}
\end{table}


\begin{table}[]
\caption{Comparing the effectiveness of finetuned A-GFN over A-GFN trained from scratch for dynamic range adjustment property constrained optimization finetuning.}
\label{tab:der_prop_const_opt}
\renewcommand{\arraystretch}{1.5}

    \resizebox{\textwidth}{!}{
    \begin{tabular}{|c|c|c|c|c|c|ccccc|c|c|c|c|}
\hline
\textbf{Task}                                                               & \textbf{\begin{tabular}[c]{@{}c@{}}Adjusted \\ range\end{tabular}} & \textbf{Method}                                                      & \textbf{N\_modes} & \textbf{Diversity} & \textbf{Success \%} & \multicolumn{5}{c|}{\textbf{L1-dist ($\downarrow$)}}                                                                                                                & \textbf{Uniqueness} & \textbf{Novelty} & \textbf{Validity} & \textbf{\begin{tabular}[c]{@{}c@{}}Time\\ (GPU-hours)\end{tabular}} \\ \hline
                                                                            &                                                                    &                                                                      &                   &                    &                     & \multicolumn{1}{c|}{TPSA} & \multicolumn{1}{c|}{\begin{tabular}[c]{@{}c@{}}Num\\ Rings\end{tabular}} & \multicolumn{1}{c|}{SAS}  & \multicolumn{1}{c|}{QED}  & Task &                     &                  &                   &                                                                     \\ \hline
\multirow{6}{*}{\begin{tabular}[c]{@{}c@{}}Molecular\\ Weight\end{tabular}} & \multirow{3}{*}{TPSA $\in [100,120]$}                              & \begin{tabular}[c]{@{}c@{}}A-GFN\\ Task Trained\end{tabular}         & 0                 & \textbf{0.94}      & 16.80               & \multicolumn{1}{c|}{1.71} & \multicolumn{1}{c|}{1.5}                                                 & \multicolumn{1}{c|}{1.87} & \multicolumn{1}{c|}{2.48} & 0.21 & 0.88                & 1.0              & 1.0               & \multirow{2}{*}{16.0}                                               \\ \cline{3-14}
                                                                            &                                                                    & \begin{tabular}[c]{@{}c@{}}A-GFN\\ Finetuned\end{tabular}            & \textbf{873}      & 0.79               & \textbf{51.70}      & \multicolumn{1}{c|}{0.16} & \multicolumn{1}{c|}{0.34}                                                & \multicolumn{1}{c|}{0.10} & \multicolumn{1}{c|}{0.59} & 0.06 & 0.94                & 1.0              & 1.0               &                                                                     \\ \cline{3-15} 
                                                                            &                                                                    & \begin{tabular}[c]{@{}c@{}}A-GFN\\ Finetuned\\ w/ data\end{tabular}  & 401               & 0.87               & 34.70               & \multicolumn{1}{c|}{0.23} & \multicolumn{1}{c|}{0.34}                                                & \multicolumn{1}{c|}{0.36} & \multicolumn{1}{c|}{0.88} & 0.09 & \textbf{1.0}        & 1.0              & 1.0               & 16.0                                                                \\ \cline{2-15} 
                                                                            & \multirow{3}{*}{TPSA $\in [40,60]$}                                & \begin{tabular}[c]{@{}c@{}}A-GFN\\ Task Trained\end{tabular}         & 0                 & \textbf{0.93}      & 11.88               & \multicolumn{1}{c|}{0.72} & \multicolumn{1}{c|}{1.50}                                                & \multicolumn{1}{c|}{1.87} & \multicolumn{1}{c|}{2.02} & 0.20 & 0.89                & 1.0              & 1.0               & \multirow{2}{*}{16.0}                                               \\ \cline{3-14}
                                                                            &                                                                    & \begin{tabular}[c]{@{}c@{}}A-GFN\\ Finetuned\end{tabular}            & \textbf{1034}     & 0.82               & \textbf{57.06}      & \multicolumn{1}{c|}{0.10} & \multicolumn{1}{c|}{0.26}                                                & \multicolumn{1}{c|}{0.15} & \multicolumn{1}{c|}{0.66} & 0.04 & 0.91                & 1.0              & 1.0               &                                                                     \\ \cline{3-15} 
                                                                            &                                                                    & \begin{tabular}[c]{@{}c@{}}A-GFN\\ Finetuned\\ w/ data\end{tabular}  & 256               & 0.87               & 38.14               & \multicolumn{1}{c|}{0.09} & \multicolumn{1}{c|}{0.35}                                                & \multicolumn{1}{c|}{0.49} & \multicolumn{1}{c|}{0.41} & 0.09 & \textbf{1.0}        & 1.0              & 1.0               & 16.0                                                                \\ \hline
\multirow{6}{*}{logP}                                                       & \multirow{3}{*}{TPSA $\in [100,120]$}                              & \begin{tabular}[c]{@{}c@{}}A-GFN\\ Task Trained\end{tabular}         & 25                & 0.74               & 16.64               & \multicolumn{1}{c|}{0.47} & \multicolumn{1}{c|}{1.5}                                                 & \multicolumn{1}{c|}{0.62} & \multicolumn{1}{c|}{0.94} & 0.30 & 0.84                & 1.0              & 1.0               & \multirow{2}{*}{72.0}                                               \\ \cline{3-14}
                                                                            &                                                                    & \begin{tabular}[c]{@{}c@{}}A-GFN\\ Finetuned\end{tabular}            & 1                 & 0.02               & 20.08               & \multicolumn{1}{c|}{0.09} & \multicolumn{1}{c|}{1.5}                                                 & \multicolumn{1}{c|}{0.11} & \multicolumn{1}{c|}{2.17} & 0.04 & 0.01                & 1.0              & 1.0               &                                                                     \\ \cline{3-15} 
                                                                            &                                                                    & \begin{tabular}[c]{@{}c@{}}A-GFN\\ Finetuned\\ w/ data\end{tabular}  & \textbf{1309}     & \textbf{0.86}      & \textbf{35.73}      & \multicolumn{1}{c|}{0.32} & \multicolumn{1}{c|}{0.26}                                                & \multicolumn{1}{c|}{0.30} & \multicolumn{1}{c|}{1.37} & 0.26 & \textbf{1.0}        & 1.0              & 1.0               & 24.0                                                                \\ \cline{2-15} 
                                                                            & \multirow{3}{*}{TPSA $\in [40,60]$}                                & \begin{tabular}[c]{@{}c@{}}A-GFN\\ Task Trained\end{tabular}         & 0                 & 0.65               & 27.50               & \multicolumn{1}{c|}{1.00} & \multicolumn{1}{c|}{1.50}                                                & \multicolumn{1}{c|}{0.02} & \multicolumn{1}{c|}{0.86} & 0.36 & 0.35                & 1.0              & 1.0               & \multirow{2}{*}{24.0}                                               \\ \cline{3-14}
                                                                            &                                                                    & \begin{tabular}[c]{@{}c@{}}A-GFN\\ Finetuned\end{tabular}            & 257               & 0.60               & \textbf{58.0}       & \multicolumn{1}{c|}{0.03} & \multicolumn{1}{c|}{1.14}                                                & \multicolumn{1}{c|}{0.03} & \multicolumn{1}{c|}{0.87} & 0.07 & 0.17                & 1.0              & 1.0               &                                                                     \\ \cline{3-15} 
                                                                            &                                                                    & \begin{tabular}[c]{@{}c@{}}A-GFN\\ Finetuned \\ w/ data\end{tabular} & \textbf{1253}     & \textbf{0.89}      & 46.29               & \multicolumn{1}{c|}{0.13} & \multicolumn{1}{c|}{0.61}                                                & \multicolumn{1}{c|}{0.28} & \multicolumn{1}{c|}{0.50} & 0.39 & \textbf{0.91}       & 1.0              & 1.0               & 28.0                                                                \\ \hline
\end{tabular}
    }
    
\end{table}


\vspace{-0.15in}
\section{Conclusion}
\vspace{-0.08in}
In this work, we introduced Atomic GFlowNets (A-GFN), an extension of the GFlowNet framework that leverages atoms as fundamental building blocks to explore molecular space. By shifting the action space from predefined molecular fragments to individual atoms, A-GFN is able to explore a much larger chemical space, enabling the discovery of more diverse and pharmacologically relevant molecules. To address the challenges posed by the vastness of this atomic action space, we propose a pretraining strategy using datasets of drug-like molecules. This off-policy pretraining approach conditions A-GFN on informative molecular properties such as drug-likeness, topological polar surface area, and synthetic accessibility, allowing it to effectively explore regions of chemical space that are more likely to yield viable drug candidates.

Our experimental results demonstrate that pretraining A-GFN with these expert trajectories leads to improved diversity and novelty in the generated molecules. Furthermore, we show that fine-tuning A-GFN for specific property optimization tasks offers significant computational efficiency compared to training from scratch. However, akin to challenges in fine-tuning RL models \citep{wolczyk2024fine}, A-GFN is prone to catastrophic forgetting of pre-trained knowledge during extended fine-tuning. This suggests a need for methods like relative trajectory balance \citep{venkatraman2024amortizing} to regularize the fine-tuning process, ensuring the policy stays anchored to the pretrained policy while still adapting to task-specific goals. Within the context of A-GFN, an obvious future direction is to extend it to lead optimization in drug discovery, where the goal is to improve candidate molecules' binding affinity, toxicity profiles, and synthetic accessibility. 

\bibliographystyle{unsrt}  
\bibliography{references}

\section*{Appendix}
\appendix
\section{Conditionals and Rewards}\label{app:cond_and_rew}
When higher values of a  molecular property $p$ or task, within desired range  $c_p = (c_{low}, c_{high})$ are preferred, $preference\_direction$ is set to $\Vec{d}>0$ and the corresponding reward is as defined in eq.\ref{eq:rew_d_pos}. Similarly, when lower values are preferred,

\begin{equation}\label{eq:rew_d_neg}
   R_p(x | c_p, \Vec{d}<0) = \text{reward}(p_x\mid c_p, \Vec{d}<0) =
\begin{cases}
\exp \left( -\frac{(\text{c$_{low}$} - p_x)}{\lambda} \right) & \text{if } p_x < \text{c$_{low}$} \\
0.5*\exp \left( -\frac{(p_x - \text{c$_{high}$})}{\lambda} \right) & \text{if } p_x > \text{c$_{high}$} \\
\frac{-0.5*(p_x - \text{c$_{low}$})}{(\text{c$_{high}$} - \text{c$_{low}$})} + 1 & \text{otherwise}
\end{cases}
\end{equation}

\vspace{5mm}

When there is no preference i.e. $\Vec{d}=0$, the reward is defined as 
\begin{equation}\label{eq:rew_d_zero}
   R_p(x | c_p, \Vec{d}<0) = \text{reward}(p_x\mid c_p, \Vec{d}<0) =
\begin{cases}
\exp \left( -\frac{(\text{c$_{low}$} - p_x)}{\lambda} \right) & \text{if } p_x < \text{c$_{low}$} \\
\exp \left( -\frac{(p_x - \text{c$_{high}$})}{\lambda} \right) & \text{if } p_x > \text{c$_{high}$} \\
 1 & \text{otherwise}
\end{cases}
\end{equation}
The rate parameter $\lambda$ is set for each property individually- $\lambda_{QED}$=$\lambda_{SAS}$=$\lambda_{NumRings}=1$, and $\lambda_{TPSA}$=20.

\begin{table}[h]

\renewcommand{\arraystretch}{1.5}

\caption{Task desired property ranges for different fine-tuning objectives. The QED, SAS, and Num Rings conditional ranges are [0.65, 0.8, 0], [1,3,0], and [1,3,1] respectively across all experiments where the ordering is [c$_{low}$,c$_{high}$,$\Vec{d}$]}
\label{tab:task_des_range}
\resizebox{\textwidth}{!}{
\begin{tabular}{|c|c|c|c|}
\hline
\textbf{Task}            & \textbf{Finetuning Objective}                            & \textbf{TPSA}   & \textbf{Task Range} \\ \hline
\multirow{4}{*}{Mol.Wt.} & Property   Optimization                                  & {[}60,100,0{]}  & {[}100,800,-1{]}    \\ \cline{2-4} 
                         & Preserved Property   Constrained Optimization            & {[}60,100,0{]}  & {[}302,800,-1{]}    \\ \cline{2-4} 
                         & \multirow{2}{*}{DRA Property Constrained   Optimization} & {[}100,120,0{]} & {[}340,800,-1{]}    \\ \cline{3-4} 
                         &                                                          & {[}40,60,0{]}   & {[}300,800,-1{]}    \\ \hline
\multirow{4}{*}{logP}    & Property   Optimization                                  & {[}60,100,0{]}  & {[}-5,6,-1{]}       \\ \cline{2-4} 
                         & Preserved Property   Constrained Optimization            & {[}60,100, 0{]} & {[}1.65,5,-1{]}     \\ \cline{2-4} 
                         & \multirow{2}{*}{DRA Property Constrained   Optimization} & {[}100,120,0{]} & {[}1.5,5,-1{]}      \\ \cline{3-4} 
                         &                                                          & {[}40,60,0{]}   & {[}2.4,5,-1{]}      \\ \hline
\multirow{4}{*}{LD50}    & Property Optimization                                  & {[}60,100,0{]}   & {[}2,6,-1{]}      \\ \cline{2-4} 
                         & Preserved Property Constrained Optimization            & {[}60,100,0{]}  & {[}2,6,-1{]}      \\ \cline{2-4} 
                         & \multirow{2}{*}{DRA Property Constrained Optimization} & {[}100,120,0{]}  & {[}1.68, 4.4,-1{]} \\ \cline{3-4} 
                         &                                                        & {[}40,60,0{]}    & {[}2,6,-1{]}      \\ \hline

\end{tabular}
}
\end{table}

\begin{table}[]
\caption{Comparing the effectiveness of finetuned atomic-GFlowNet over atomic-GFlowNet trained from scratch for
TPSA property targeting objective.
}
\label{tab:prop_tgt_opt}
\renewcommand{\arraystretch}{1.5}

\resizebox{\textwidth}{!}{
\begin{tabular}{|l|l|l|l|l|llll|l|l|l|l|}
\hline
\textbf{\begin{tabular}[c]{@{}l@{}}Adjusted Range\end{tabular}}       & \textbf{Method}                                                   & \textbf{N\_modes} & \textbf{Diversity} & \textbf{Success \%} & \multicolumn{4}{c|}{\textbf{L1-dist ($\downarrow$)}}                                                                                      & \textbf{Uniqueness} & \textbf{Novelty} & \textbf{Validity} & \textbf{\begin{tabular}[c]{@{}l@{}}Time\\ (GPU-hours)\end{tabular}} \\ \hline
                                                                         &                                                                   &                   &                    &                     & \multicolumn{1}{l|}{TPSA} & \multicolumn{1}{l|}{\begin{tabular}[c]{@{}l@{}}Num \\ Rings\end{tabular}} & \multicolumn{1}{l|}{SAS}  & QED  &                     &                  &                   &                                                                     \\ \hline
\multirow{2}{*}{\begin{tabular}[c]{@{}l@{}}TPSA $\in$ [100,120]\end{tabular}} & \begin{tabular}[c]{@{}l@{}}A-GFN\\ Task Trained\end{tabular}    & 0                 & 0.01               & 24.71               & \multicolumn{1}{l|}{0.01} & \multicolumn{1}{l|}{1.5}                                                  & \multicolumn{1}{l|}{0.62} & 2.40 & 0.01                & 1.0              & 1.0               & \multirow{2}{*}{73.0}                                               \\ \cline{2-12}
                                                                         & \begin{tabular}[c]{@{}l@{}}A-GFN\\ Finetuned\end{tabular}          & 254               & 0.60               & 46.30               & \multicolumn{1}{l|}{0.07} & \multicolumn{1}{l|}{1.50}                                                 & \multicolumn{1}{l|}{0.01} & 1.23 & 0.50                & 1.0              & 1.0               &                                                                     \\ \hline
\multirow{2}{*}{TPSA $\in$ [40,60]}                                              & \begin{tabular}[c]{@{}l@{}}A-GFN\\Task Trained\end{tabular} & 0                 & 0.39               & 46.83               & \multicolumn{1}{l|}{0.01} & \multicolumn{1}{l|}{1.5}                                                  & \multicolumn{1}{l|}{0.00} & 1.35 & 0.04                & 1.0              & 1.0               & \multirow{2}{*}{10.0}                                               \\ \cline{2-12}
                                                                         & \begin{tabular}[c]{@{}l@{}}A-GFN\\ Finetuned\end{tabular}          & 616               & 0.83               & 55.60               & \multicolumn{1}{l|}{0.10} & \multicolumn{1}{l|}{0.22}                                                 & \multicolumn{1}{l|}{0.13} & 0.55 & 0.80                & 1.0              & 1.0               &                                                                     \\ \hline
\end{tabular}
}
\end{table}

\begin{table}[h]
\caption{Property constrained optimization for toxicity (LD50) task }
\label{tab:LD50_const_prop_opt}
\renewcommand{\arraystretch}{1}

\resizebox{\textwidth}{!}{
\begin{tabular}{|c|c|c|c|c|ccccc|c|c|c|c|}
\hline
\textbf{Adjusted range}               & \textbf{Method}                                             & \textbf{N\_modes} & \textbf{Diversity} & \textbf{Success \%} & \multicolumn{5}{c|}{\textbf{L1-dist ($\downarrow$)}}                                                                                                                & \textbf{Uniqueness} & \textbf{Novelty} & \textbf{Validity} & \textbf{\begin{tabular}[c]{@{}c@{}}Time\\ (GPU-hours)\end{tabular}} \\ \hline
                                      &                                                             &                   &                    &                     & \multicolumn{1}{c|}{TPSA} & \multicolumn{1}{c|}{\begin{tabular}[c]{@{}c@{}}Num\\ Rings\end{tabular}} & \multicolumn{1}{c|}{SAS}  & \multicolumn{1}{c|}{QED}  & Task &                     &                  &                   &                                                                     \\ \hline
\multirow{2}{*}{Preserved}            & \begin{tabular}[c]{@{}c@{}}A-GFN\\ tasktrained\end{tabular} &                  0 &        0.37            &      57.91               & \multicolumn{1}{c|}{0.02}     & \multicolumn{1}{c|}{1.2}                                                    & \multicolumn{1}{c|}{0.01}     & \multicolumn{1}{c|}{1.56}     &   0.74   &              0.24      &          0.24      &        1.0           &           24.0                                                          \\ \cline{2-14} 
                                      & \begin{tabular}[c]{@{}c@{}}A-GFN\\ finetuned\end{tabular}   &             682      & 0.61               & 55.16               & \multicolumn{1}{c|}{0.02} & \multicolumn{1}{c|}{0.30}                                                & \multicolumn{1}{c|}{0.11} & \multicolumn{1}{c|}{3.68} & 0.49 & 0.99                & 1.0              & 1.0               &         24.0                                                            \\ \hline
\multirow{2}{*}{TPSA $\in [100,120]$} & \begin{tabular}[c]{@{}c@{}}A-GFN\\ tasktrained\end{tabular} &                 0  &          0.75          &         28.95            & \multicolumn{1}{c|}{0.20}     & \multicolumn{1}{c|}{1.49}                                                    & \multicolumn{1}{c|}{0.40}     & \multicolumn{1}{c|}{2.789}     &  0.73    &        0.99             &        0.99         &           1.0        &         24.0                                                            \\ \cline{2-14} 
                                      & \begin{tabular}[c]{@{}c@{}}A-GFN\\ finetuned\end{tabular}   &               308    & 0.70               & 38.52               & \multicolumn{1}{c|}{0.30} & \multicolumn{1}{c|}{1.49}                                                & \multicolumn{1}{c|}{0.05} & \multicolumn{1}{c|}{4.06} & 0.67 & 0.96                & 1.0              & 1.0               &     24.0                                                                \\ \hline
\multirow{2}{*}{TPSA $\in [40,60]$}   & \begin{tabular}[c]{@{}c@{}}A-GFN\\ tasktrained\end{tabular} &               0    &      0.04
&          20           & \multicolumn{1}{c|}{0.48}     & \multicolumn{1}{c|}{1.5}                                                    & \multicolumn{1}{c|}{0.039}     & \multicolumn{1}{c|}{2.38}     &   0.61   &            0.002         &           0.002      &         1.0          &     24.0                                                                \\ \cline{2-14} 
                                      & \begin{tabular}[c]{@{}c@{}}A-GFN\\ finetuned\end{tabular}   &                493   & 0.63               & 47.61               & \multicolumn{1}{c|}{0.36} & \multicolumn{1}{c|}{0.09}                                                & \multicolumn{1}{c|}{0.07} & \multicolumn{1}{c|}{2.68} & 0.47 & 0.70                & 1.0              & 1.0               &       24.0                                                              \\ \hline
\end{tabular}
}
\end{table}

\newpage

\section{ Evaluation Metrics}\label{app:metrics}
In line with earlier works \citep{brown2019guacamol, you2018graph}, we employ following set of metrics to ensure a robust comparison across different generative methods. \\
\textbf{Validity}: A molecule is considered valid if it successfully passes RDKit's sanitization checks. Note that in the proposed method, masking makes this trivially 1. \\
\textbf{Diversity}: For a set of generated molecules, diversity is defined based on the Tanimoto similarity of Morgan Fingerprint representation of molecules. 
$$Diversity = 1- \frac{2}{N(N-1)} \sum_{1\leq i\leq j \leq N} \frac{|M_i \cap M_j|}{|M_i \cup M_j|} $$\\ 
\textbf{Uniqueness}: The ratio of distinct canonicalized SMILES strings (without stereochemistry) to the total number of generated molecules, after filtering out duplicates and invalid structures.\\
\textbf{Novelty}: The ratio of unique generated molecules, that are not present in the pretraining ZINC dataset, to the total number of generated molecules.\\
\textbf{Time}: For each task, a fixed time budget is allotted to A-GFN for fine-tuning and task-training. Other baseline methods are allowed to run for time needed to run their default configuration.\\
\textbf{N\_Modes}: This metric quantifies the number of distinct, high-reward molecular modes identified by generative model. A mode is defined as a molecule whose reward exceeds a threshold, typically set at the 75$^{th}$ percentile of the reward distribution achieved by the best-performing model for the task. For a molecule to be counted as a new mode, its Tanimoto similarity to any previously identified mode must be less than a specified threshold (typically 0.7), ensuring that only sufficiently diverse molecules are counted.\\
\textbf{Normalized L1-dist}: This measures how far a generated molecule's properties deviate from their target ranges. For each property, the L1 distance is calculated by taking the absolute difference between the generated property value and the 10$^{th}$ percentile value in desired range, and then normalized by the respective property range to ensure comparability across properties. This allows us to assess how well the generated molecules meet multi-objective constraints in a unified metric.\\
\textbf{Success Percent}: For a set of $N$ generated graphs $\{G\}_{i=1}^N$ and conditionals $C_{task}$ for given task, we define success percentage as 
\[
S_{\text{task}} = \frac{1}{N} \sum_{i=1}^N \left( \frac{1}{|\mathcal{C}_{\text{task}}|} \sum_{c \in \mathcal{C}_{\text{task}}} \mathbb{I}_c(G_i) \right) \times 100
\]

where, $\mathbb{I}_c(G)$ counts the molecules within desired conditional ranges and defined as  

\[
\mathbb{I}_c(G) = 
\begin{cases} 
\mathbb{I}\left(|c(G) - c_{\text{low}}| \leq 0.1 \cdot |c_{\text{low}}|\right), & \text{if } \text{preference\_direction}(c) < 0 \\ 
\mathbb{I}\left(|c(G) - c_{\text{high}}| \leq 0.1 \cdot |c_{\text{high}}|\right), & \text{if } \text{preference\_direction}(c) > 0 \\ 
\mathbb{I}\left(c_{\text{low}} \leq c(G) \leq c_{\text{high}}\right), & \text{if } \text{preference\_direction}(c) = 0 
\end{cases}
\]

where, for any considered task, $c(G)$ is the the value of the conditional property $c$ for the molecular graph $G$, $c_{low}$  and $c_{high}$  are the lower and upper bounds of $c$, $preference\_direction(c)$ indicates whether lower values ($<0$), higher values ($>0$), or values within a range ($=0$) are preferred.

\begin{figure}[h]
\includegraphics[width=\textwidth,height=\textheight,keepaspectratio]{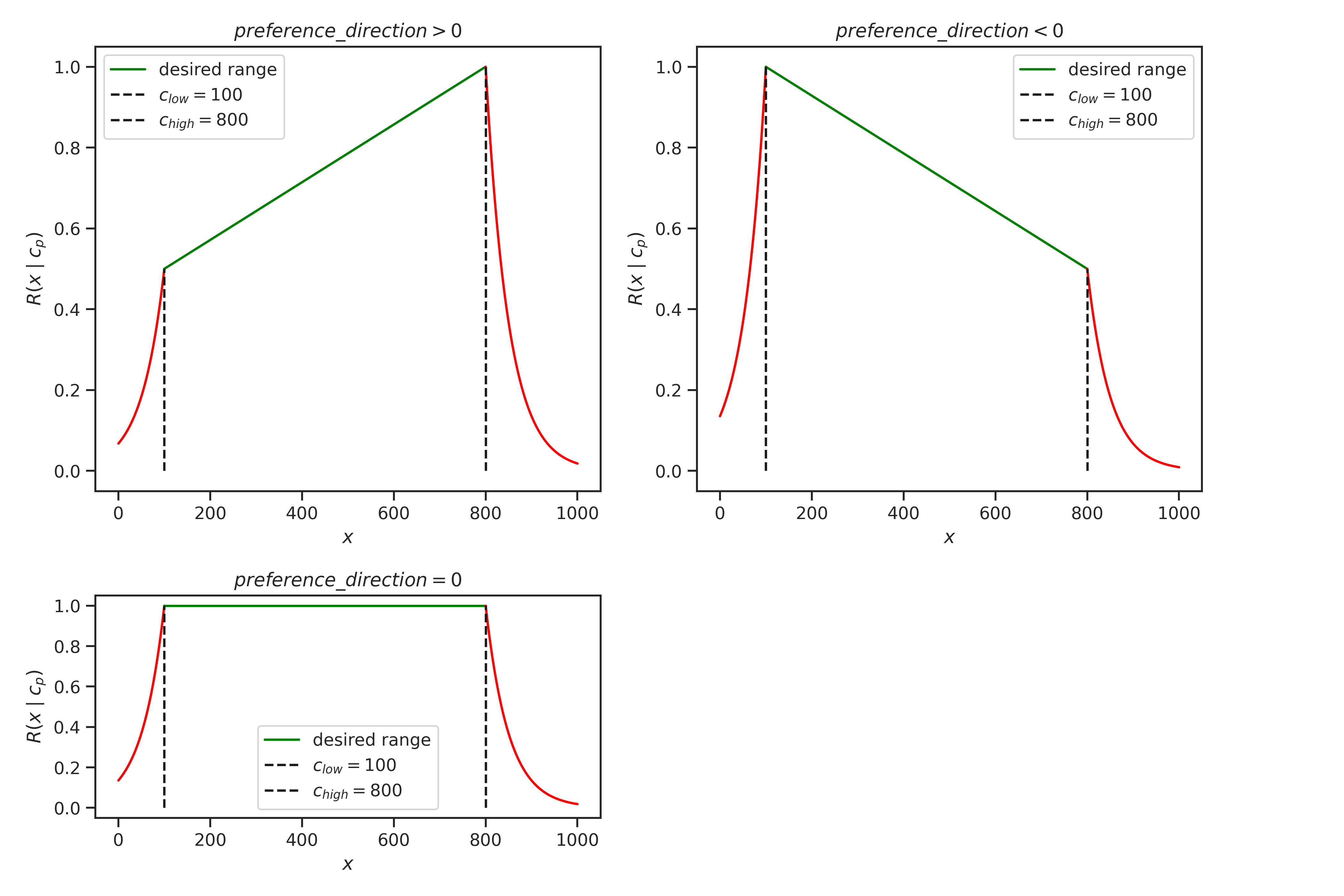} 
  \centering
  \caption{Visual representation of reward as a function of $preference\_direction$.}
  \label{fig:pref_dir}
\end{figure}


\section{Experiments}

\begin{table}[]
\centering
\caption{Pretraining and finetuning setups and corresponding hyperparameters} \label{tab:hps}
\renewcommand{\arraystretch}{1.2}
\begin{tabular}{|c|ccc|}
\hline
\textbf{Hyperparameter}        & \multicolumn{1}{c|}{\textbf{\begin{tabular}[c]{@{}c@{}}Pretraining\\ Values\end{tabular}}} & \multicolumn{1}{c|}{\textbf{\begin{tabular}[c]{@{}c@{}}Finetuning\\ Values\end{tabular}}} & \textbf{\begin{tabular}[c]{@{}c@{}}FT w/data\\ Values\end{tabular}} \\ \hline
max\_num\_iter                 & \multicolumn{3}{c|}{500,000}                                                                                                                                                                                                                                 \\ \hline
bootstrap\_own\_reward         & \multicolumn{3}{c|}{FALSE}                                                                                                                                                                                                                                   \\ \hline
random\_seed                   & \multicolumn{3}{c|}{1,428,570}                                                                                                                                                                                                                               \\ \hline
beta                           & \multicolumn{1}{c|}{96}                                                                    & \multicolumn{2}{c|}{64}                                                                                                                                         \\ \hline
OOB\_percent                   & \multicolumn{3}{c|}{0.1}                                                                                                                                                                                                                                     \\ \hline
zinc\_rad\_scale ($\lambda$)   & \multicolumn{3}{c|}{1}                                                                                                                                                                                                                                       \\ \hline
gfn\_batch\_shuffle            & \multicolumn{3}{c|}{FALSE}                                                                                                                                                                                                                                   \\ \hline
reward\_aggregation            & \multicolumn{3}{c|}{mul}                                                                                                                                                                                                                                     \\ \hline
sampling\_batch\_size          & \multicolumn{1}{c|}{2048}                                                                  & \multicolumn{2}{c|}{1024}                                                                                                                                       \\ \hline
training\_batch\_size          & \multicolumn{3}{c|}{64}                                                                                                                                                                                                                                      \\ \hline
learning\_rate                 & \multicolumn{3}{c|}{1.00e-04}                                                                                                                                                                                                                                \\ \hline
online\_offline\_mix\_ratio    & \multicolumn{3}{c|}{0.5}                                                                                                                                                                                                                                     \\ \hline
num\_workers                   & \multicolumn{1}{c|}{8}                                                                     & \multicolumn{2}{c|}{2}                                                                                                                                          \\ \hline
gfn\_loss\_coeff ($\lambda_1$) & \multicolumn{1}{c|}{0.04}                                                                  & \multicolumn{1}{c|}{-}                                                                    & 0.04                                                                \\ \hline
MLE\_coeff ($\lambda_2$)       & \multicolumn{1}{c|}{20}                                                                    & \multicolumn{1}{c|}{-}                                                                    & 20                                                                  \\ \hline
num\_emb                       & \multicolumn{3}{c|}{128}                                                                                                                                                                                                                                     \\ \hline
num\_layers                    & \multicolumn{3}{c|}{8}                                                                                                                                                                                                                                       \\ \hline
num\_mlp\_layers               & \multicolumn{3}{c|}{4}                                                                                                                                                                                                                                       \\ \hline
num\_heads                     & \multicolumn{3}{c|}{2}                                                                                                                                                                                                                                       \\ \hline
i2h\_width                     & \multicolumn{3}{c|}{1}                                                                                                                                                                                                                                       \\ \hline
illegal\_action\_logreward     & \multicolumn{3}{c|}{-512}                                                                                                                                                                                                                                    \\ \hline
reward\_loss\_multiplier       & \multicolumn{3}{c|}{1}                                                                                                                                                                                                                                       \\ \hline
weight\_decay                  & \multicolumn{3}{c|}{1.00e-08}                                                                                                                                                                                                                                \\ \hline
num\_data\_loader\_workers     & \multicolumn{3}{c|}{8}                                                                                                                                                                                                                                       \\ \hline
momentum                       & \multicolumn{3}{c|}{0.9}                                                                                                                                                                                                                                     \\ \hline
adam\_eps                      & \multicolumn{3}{c|}{1.00e-08}                                                                                                                                                                                                                                \\ \hline
lr\_decay                      & \multicolumn{3}{c|}{20,000}                                                                                                                                                                                                                                  \\ \hline
Z\_lr\_decay                   & \multicolumn{3}{c|}{20,000}                                                                                                                                                                                                                                  \\ \hline
clip\_grad\_type               & \multicolumn{3}{c|}{norm}                                                                                                                                                                                                                                    \\ \hline
clip\_grad\_param              & \multicolumn{3}{c|}{10}                                                                                                                                                                                                                                      \\ \hline
random\_action\_prob           & \multicolumn{3}{c|}{0.001}                                                                                                                                                                                                                                   \\ \hline
random\_stop\_prob             & \multicolumn{3}{c|}{0.001}                                                                                                                                                                                                                                   \\ \hline
num\_back\_steps\_max          & \multicolumn{3}{c|}{25}                                                                                                                                                                                                                                      \\ \hline
max\_traj\_len                 & \multicolumn{3}{c|}{40}                                                                                                                                                                                                                                      \\ \hline
max\_nodes                     & \multicolumn{3}{c|}{45}                                                                                                                                                                                                                                      \\ \hline
max\_edges                     & \multicolumn{3}{c|}{50}                                                                                                                                                                                                                                      \\ \hline
tb\_p\_b\_is\_parameterized    & \multicolumn{3}{c|}{TRUE}                                                                                                                                                                                                                                    \\ \hline
num\_thermometer\_dim          & \multicolumn{3}{c|}{16}                                                                                                                                                                                                                                      \\ \hline
sample\_temp                   & \multicolumn{3}{c|}{1}                                                                                                                                                                                                                                       \\ \hline
checkpoint\_every              & \multicolumn{1}{c|}{1,000}                                                                 & \multicolumn{2}{c|}{500}                                                                                                                                        \\ \hline
Z\_learning\_rate              & \multicolumn{3}{c|}{1e-3}                                                                                                                                                                                                                                    \\ \hline
\end{tabular}
\end{table}

\end{document}